\documentclass[conference]{IEEEtran}
\IEEEoverridecommandlockouts
\usepackage{amsmath,amssymb,amsfonts}
\usepackage{graphicx}
\usepackage{textcomp}
\usepackage{xcolor}
\usepackage{comment}
\usepackage{float}
\usepackage{algorithm}
\usepackage{algpseudocode}
\usepackage{amsmath}
\usepackage{amssymb}
\usepackage{multirow}
\usepackage{mathtools}

\usepackage{biblatex} 
\usepackage{titlesec}
\titlespacing*{\section}{0pt}{0.5\baselineskip}{0.3\baselineskip}
\titlespacing*{\subsection}{0pt}{0.4\baselineskip}{0.25\baselineskip}
\titlespacing*{\subparagraph}{0pt}{0.25\baselineskip}{0.25\baselineskip}

\def\BibTeX{{\rm B\kern-.05em{\sc i\kern-.025em b}\kern-.08em
    T\kern-.1667em\lower.7ex\hbox{E}\kern-.125emX}}

\makeatletter

\makeatletter

\begin{document}
\title{
Uncertainty-Gated Region-Level Retrieval for Robust Semantic Segmentation
}

\author{
\IEEEauthorblockN{Shreshth Rajan}
\IEEEauthorblockA{
Harvard University \\
Cambridge, MA, USA \\
shreshthrajan@college.harvard.edu}
\and
\IEEEauthorblockN{Raymond Liu}
\IEEEauthorblockA{
Harvard University \\
Cambridge, MA, USA \\
liur@g.harvard.edu}
}

\maketitle

\begin{abstract}
Semantic segmentation of outdoor street scenes plays a key role in applications such as autonomous driving, mobile robotics, and assistive technology for visually-impaired pedestrians. For these applications, accurately distinguishing between key surfaces and objects such as roads, sidewalks, vehicles, and pedestrians is essential for maintaining safety and minimizing risks. Semantic segmentation must be robust to different environments, lighting and weather conditions, and sensor noise, while being performed in real-time. We propose a region-level, uncertainty-gated retrieval mechanism that improves segmentation accuracy and calibration under domain shift. Our best method achieves an 11.3\% increase in mean intersection-over-union while reducing retrieval cost by 87.5\%, retrieving for only 12.5\% of regions compared to 100\% for always-on baseline.
\end{abstract}

\section{Introduction and Related Work}

Semantic segmentation is a computer vision task that assigns a class label to every pixel in an image. In outdoor street scenes, semantic segmentation is used in applications such as autonomous driving, mobile robotics, and visual assistants for visually-impaired pedestrians to detect key surfaces and objects such as roads, sidewalks, vehicles, and pedestrians. Performing semantic segmentation accurately and in real-time is crucial in these applications where inaccurate or delayed predictions can pose serious safety risks.

Semantic segmentation of outdoor street scenes suffers from multiple degrees of domain shift: different locations, day/night conditions, weather conditions, and various image artifacts can all severely degrade segmentation model performance \cite{kamann2020, michaelis2019}. Datasets containing high-quality segmentation labels of outdoor street scenes are typically largely comprised of images taken during the daytime with clear weather conditions \cite{cityscapes, mapillary}, or are limited to a few cities \cite{sanpo}. Segmentation models trained on images from one or more of these datasets may not generalize well to other domains.

\subsection{Retrieval-Augmented Segmentation}

Memory-based retrieval has been explored for semantic segmentation under domain shift. Pin the Memory \cite{pin} improved segmentation of urban scenes but required retraining with memory guidance. More recent work on few-shot medical image segmentation \cite{medical} retrieves similar samples at the image level using DINOv2 features and SAM2's memory attention, achieving strong performance without retraining. However, image-level retrieval is computationally expensive for dense prediction tasks.

\subsection{Uncertainty Estimation}

Test-time augmentation has been used to estimate uncertainty by measuring prediction variability across augmented inputs. Mutual information, which isolates epistemic uncertainty from aleatoric uncertainty, has shown promise for identifying model uncertainty that can be reduced through additional information \cite{uncertaintyuq}. Expected Pairwise KL Divergence (EPKL) measures disagreement between ensemble predictions and has achieved strong correlation ($r>0.9$) with segmentation quality on medical imaging datasets \cite{epklpaper}.

\subsection{Foundation Models}

DINOv2 \cite{dinov2} is a self-supervised vision transformer trained on 142M images that produces robust semantic features. Region-level representations \cite{regionrep} using DINOv2 have shown to be effective for retrieval tasks while being computationally efficient compared to image-level approaches.

\begin{figure}[!t]
    \centering
    \includegraphics[width=\linewidth]{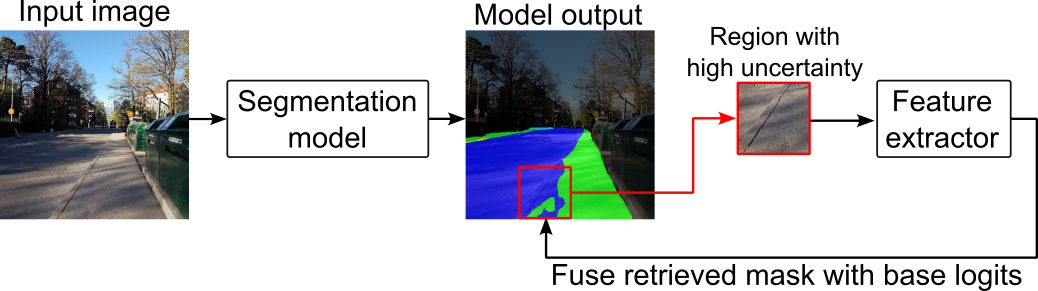}
    \caption{Retrieval is used on regions of high uncertainty to improve the robustness of outputs from a lightweight segmentation model.}
    \label{retrieval}
\end{figure}

We propose a selective uncertainty-gated retrieval mechanism for domain adaptation that improves segmentation accuracy and calibration under domain shift without retraining (Fig.~\ref{retrieval}). Given a region of interest in an image, we retrieve similar regions from a memory bank and fuse their corresponding probability maps with the base model logits. Similarity is determined using embeddings from DINOv2 \cite{dinov2}. Retrieval is only done in regions with high uncertainty, allowing the model to adaptively refine its predictions without excessive latency.

\section{Initial Approach}

We implemented region-level uncertainty detection with SegFormer-B0~\cite{segformer}, a powerful and lightweight segmentation model with 3.7M parameters. The SegFormer-B0 model was fine-tuned on the Cityscapes dataset~\cite{cityscapes} containing vehicle-egocentric images and corresponding segmentation labels. For evaluation, we used the Cityscapes validation set.

Uncertainty is measured using test-time augmentation. We generate five total predictions: one from the original image and four from augmented versions created using horizontal flip, rescaling (scales 0.9 and 1.1), and color jitter (brightness 0.1, contrast 0.1, saturation 0.05, hue 0.02). Initially, we computed uncertainty using pixel-wise entropy of the mean prediction. We later refined this to use mutual information, which isolates epistemic uncertainty by measuring the information gain from observing model predictions.

Uncertain regions are determined by finding connected components of pixels with high uncertainty. Given an uncertainty map for an image, pixels at or above the 75th percentile of uncertainty are considered to have high uncertainty. We find connected components of high uncertainty pixels containing at least 100 pixels, then get the bounding box of each connected components to use as an uncertain region.

Global and region-level features are extracted using DINOv2 ViT-B/14. The global features for an image are the 768-dimensional output embedding vector for the DINOv2 CLS token. The features for a region are the result of applying Region of Interest Align to each of the 768-dim patch embeddings for the patches that the region's bounding box overlaps with.

We use a memory bank for similarity matching and retrieval of region-level probability maps. The memory bank contains information from 200 images from the first portion of the Cityscapes validation set (images 0-199), while we evaluate on a disjoint set of 100 validation images (217-316) with motion blur corruption. Due to storage constraints, we use a validation set split rather than the full training set for the memory bank. For each image, the global and region-level features are extracted using DINOv2 and saved in the memory bank. The corresponding region-level ground truth class labels are also saved and later converted to probability maps during fusion with base model logits. Only the top 25\% most confident (lowest uncertainty) regions are saved in the memory bank.

During base model inference, the top 25\% of uncertain regions are matched with similar regions in the memory bank. Region matching is performed hierarchically. First, global image features are matched using cosine similarity to find the top-50 most similar images. Among these images, we then compute the top-5 most similar regions by cosine similarity. The corresponding probability maps for these regions are fused with the base model logits using a similarity-based weighting mechanism that assigns higher weights to regions with higher cosine similarities.

\section{Initial Results}

\begin{figure}[!t]
    \centering
    \includegraphics[width=\linewidth]{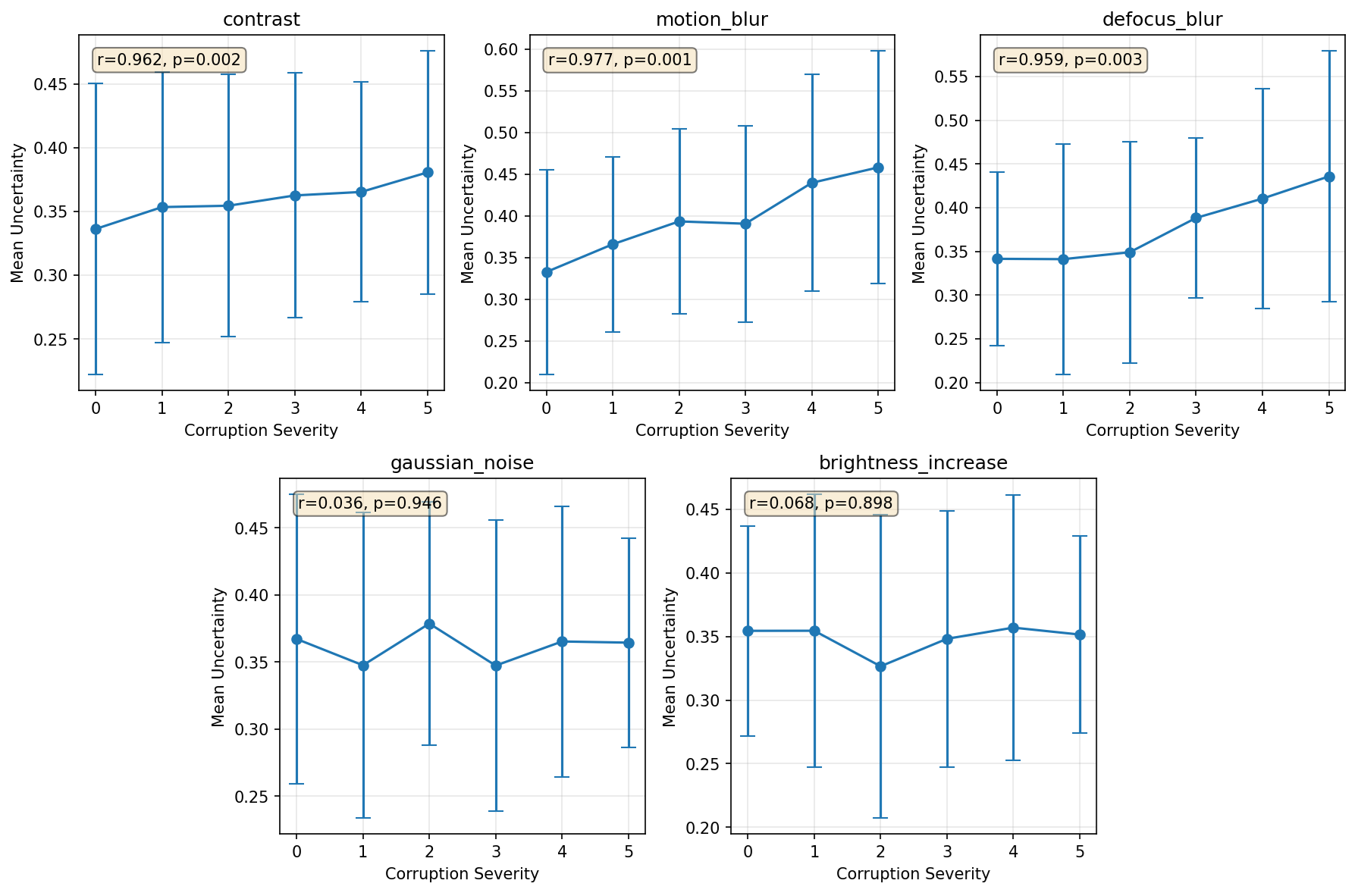}
    \caption{Uncertainty increases with severity of certain data augmentations.}
    \label{calibration}
\end{figure}

We performed all validation experiments on 100 images from the Cityscapes validation set. Baseline performance was measured by the mean intersection-over-union (IoU) of the combined SegFormer predictions across augmented images to simulate performance under domain shift. Our initial analysis showed that data augmentation increased uncertainty, demonstrating calibration in our uncertainty measurements (Fig.~\ref{calibration}).

\begin{figure}[!t]
    \centering
    \includegraphics[width=\linewidth]{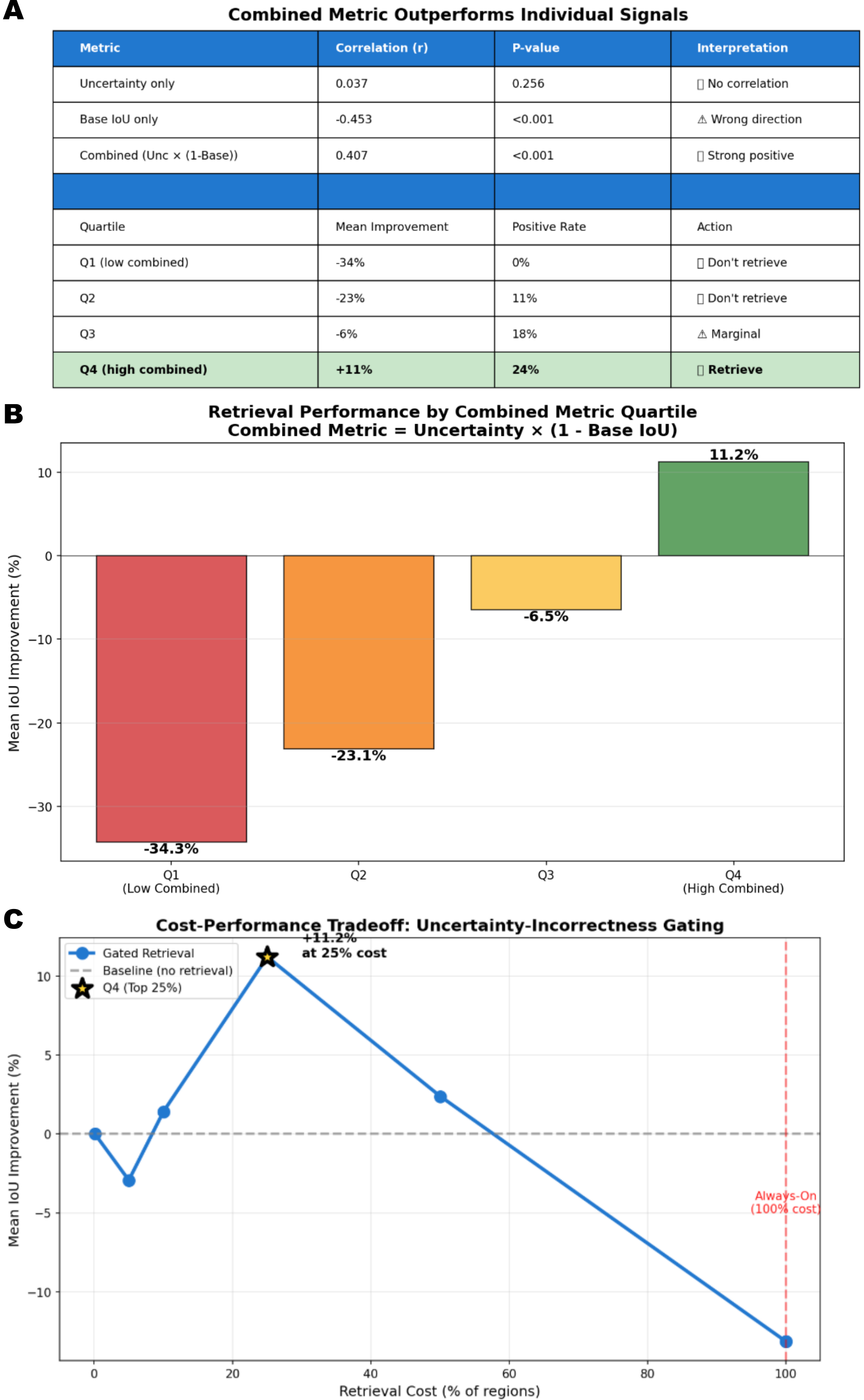}
    \caption{\textbf{(A)} Combining uncertainty and performance outperforms individual signals. \textbf{(B,C)} Gating on the top 25\% of regions based on the combined metric achieves the best cost-performance tradeoff.}
    \label{1-combined}
\end{figure}

However, our experiments revealed that uncertainty did not predict retrieval benefit. Region-level uncertainty was uncorrelated with changes in IoU ($r=0.04$, $p=0.26$). Directly using the IoU of the baseline predictions was negatively correlated with retrieval benefit ($r=-0.45$, $p<0.001$). We discovered that a combined metric of \textbf{uncertainty$\times$(1~-~base\_iou)} was positively correlated with performance ($r=0.41$, $p<0.001$) (Fig.~\ref{1-combined}A). Improper gating can hurt performance; gating on the top 25\% of regions based on the combined metric achieved the best cost-performance tradeoff, with an +11\% improvement in IoU and 75\% lower retrieval cost compared to always-on retrieval (Fig.~\ref{1-combined}B,C).

\section{Revised Approach and Results}

Although the combined metric can improve model accuracy, it is unrealistic for normal use because it relies directly on knowledge of ground truth labels. We investigated alternative metrics that do not require ground truth.

\begin{figure}[!t]
    \centering
    \includegraphics[width=\linewidth]{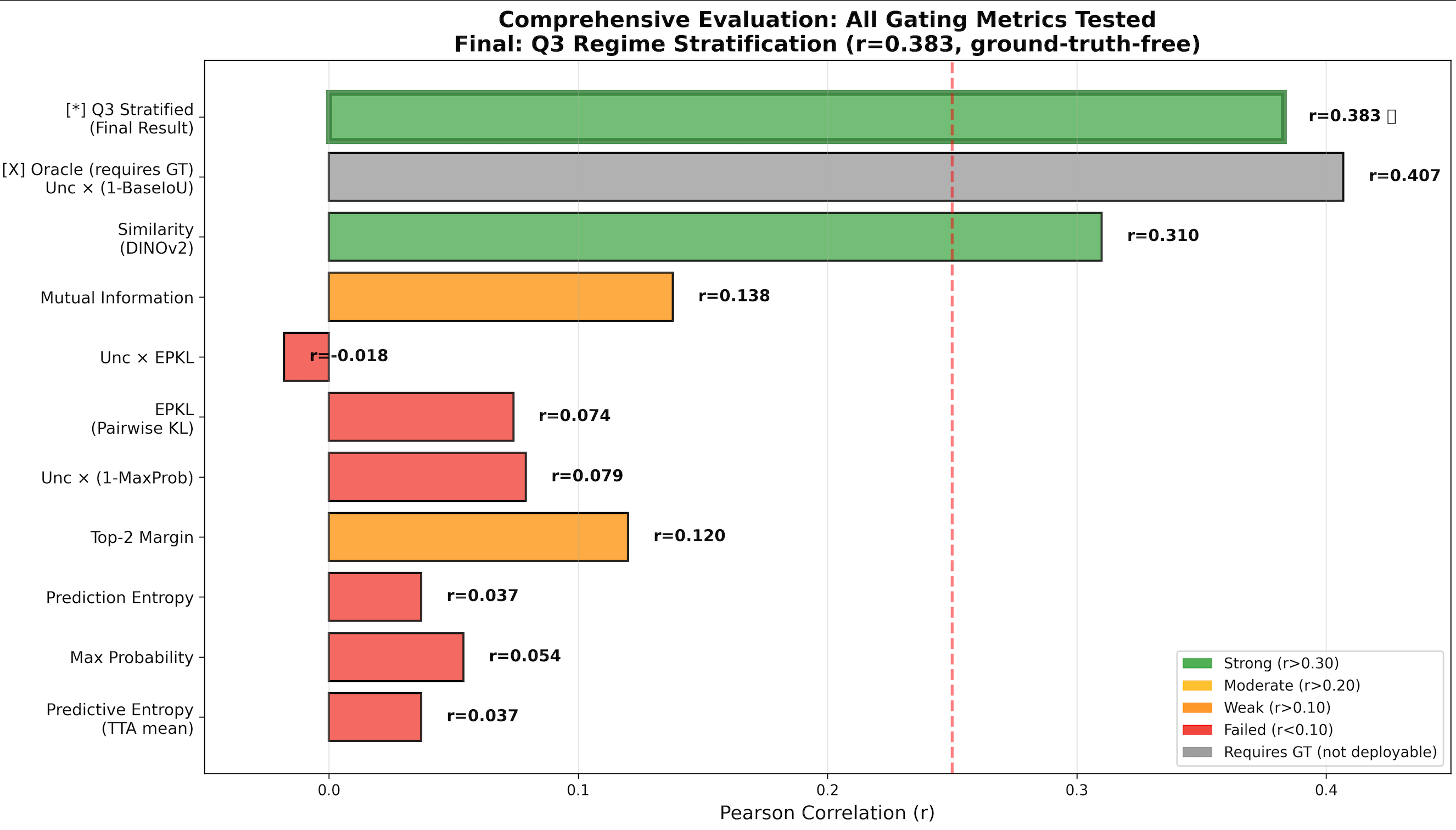}
    \caption{Overview of tested gating methods.}
    \label{comprehensive}
\end{figure}

We initially tried a variety of alternative approaches, including maximum prediction probability ($r=0.05$), prediction entropy ($r=0.04$), top-2 class margin ($r=0.12$), uncertainty combined with inverted maximum probability ($r=0.08$), Expected Pairwise KL Divergence (EPKL, $r=0.07$), and uncertainty combined with EPKL ($r=-0.02$). All of these metrics showed weak or negligible correlation with retrieval benefit (Fig.~\ref{comprehensive}) 

Eventually, one approach that showed promise was to use the mutual information between the image augmentations to calculate uncertainty, rather than using the class entropy. We found that using mutual information during test-time augmentation showed a weak positive correlation with IoU improvement ($r=0.14, p<0.001$) (Fig.~\ref{comprehensive}). We stratified all regions by mutual information into quartiles and measured how well similarity predicted retrieval benefit within each quartile. The third quartile of regions (MI $\in [0.69, 0.76]$) showed the strongest similarity correlation ($r=0.38$), while the fourth quartile (very high MI) showed much weaker correlation ($r=0.21$). This reveals that moderate epistemic uncertainty defines an optimal retrieval regime, while very low MI regions are typically correct and very high MI regions represent ambiguous cases that are difficult to improve even with retrieval

Independent of uncertainty metrics, we found that retrieval quality itself was a strong predictor of benefit. DINOv2 semantic similarity (cosine similarity between region features) showed the strongest correlation with IoU improvement among all ground-truth-free metrics ($r=0.31, p<0.001$) (Fig.~\ref{comprehensive}). This indicates that retrieval helps when the memory bank contains semantically similar regions, regardless of model uncertainty. However, similarity alone cannot identify which regions need retrieval in the first place.

\begin{figure}[!t]
    \centering
    \includegraphics[width=\linewidth]{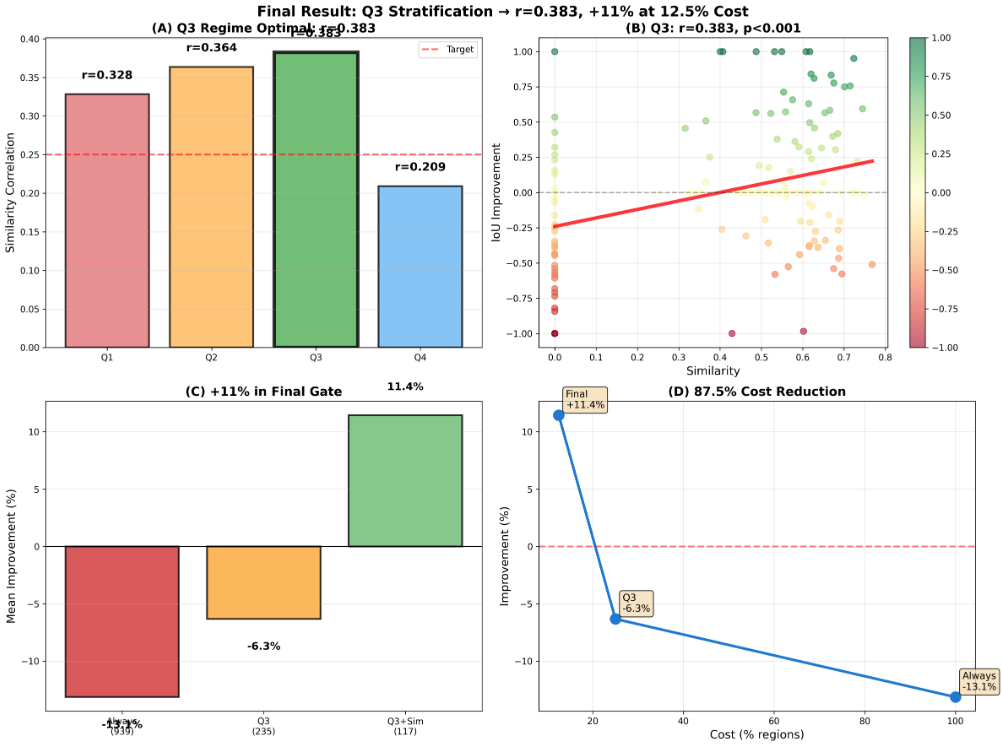}
    \caption{Final results for refined uncertainty and similarity gating.}
    \label{final}
\end{figure}

As shown in Fig.~\ref{final}A, the third quartile demonstrates that moderate epistemic uncertainty creates the optimal regime where similarity-based gating is most effective. Building on this discovery, we implemented a two-stage gating mechanism: first filtering to Q3 by mutual information, then ranking by similarity within that regime. Regions that pass the uncertainty gate are then matched with regions in the memory bank by similarity. Uncertainty-gated region similarity showed a positive correlation with IoU improvement ($r=0.38$, $p<0.001$) (Fig.~\ref{final}B); the top 50\% of regions by similarity are kept for retrieval while the bottom 50\% are discarded. This two-stage gating achieves an 11.4\% improvement in IoU on targeted regions (117/939, 12.5\%) while reducing retrieval cost by 87.5\% compared to always-on baseline. In contrast, always-on retrieval (100\% of regions) results in a 13.1\% reduction in IoU, demonstrating that indiscriminate retrieval degrades performance (Fig.~\ref{final}C,D).

\subsection{Region-Level Performance Analysis}

Analysis of retrieval benefit by base prediction quality reveals distinct patterns. Regions where the base model performs poorly (base IoU $<$ 0.2) show positive mean improvement (+10.5\%) with 22\% success rate, indicating retrieval can correct major errors. Conversely, regions with high base accuracy (IoU $>$ 0.8) show mean degradation of -32.6\% with only 2.6\% success rate. This demonstrates that retrieval primarily helps when base predictions are substantially wrong, while high-confidence correct predictions are degraded by fusion with retrieved information.

\subsection{Failure Mode Analysis}

We analyzed 270 cases where retrieval caused significant harm (IoU degradation $>$ 20\%). These failures predominantly occur in regions where base predictions were already accurate (mean base IoU = 0.67) but retrieval similarity was low (mean similarity = 0.24). In these cases, hierarchical filtering failed to find semantically similar regions, leading to fusion with contextually mismatched retrieved masks. This reveals a fundamental limitation: when the memory bank lacks appropriate matches, retrieval should be avoided entirely rather than forced. The similarity threshold in our two-stage gating addresses this by filtering out low-quality matches.

\section{Conclusion}

We introduce a method for uncertainty- and similarity-gated retrieval that improves semantic segmentation accuracy while minimizing retrieval cost from a memory bank. Through epistemic uncertainty stratification, we discover that moderate mutual information (Q3, MI $\in [0.69, 0.76]$) defines an optimal retrieval regime where similarity-based gating achieves $r=0.38$ correlation with benefit. Our final two-stage gating method improves IoU on synthetically-corrupted Cityscapes validation images by 11.4\% while reducing retrieval cost by 87.5\%, performing retrieval on only 12.5\% of regions compared to 100\% for always-on baseline. This demonstrates that selective retrieval based on epistemic uncertainty and semantic similarity enables both accuracy gains and computational efficiency.

Future work should evaluate with more data and across different datasets, datasets with pedestrian-egocentric images, to better model domain shift. More realistic corruptions such as dynamic lighting and weather condition simulation can also model domain shift more realistically. Our work focused on test-time augmentation, but other types of measuring uncertainty such as Monte Carlo dropout, deep ensembles, and Gaussian process probes are worth investigating. While we focus on region-level uncertainty based on connected components of uncertain pixels, it is possible to extract uncertain regions based on ViT patches and directly patch-level embeddings for similarity. Evaluation with models and encoders other than SegFormer and DINOv2 is needed to show generalizability. We used IoU to evaluate accuracy, but other segmentation performance metrics such as precision, recall, and pixel accuracy are worth evaluating as well.


\section*{Acknowledgments}

We thank David Belanger, Zi Wang, and the rest of the CS 2823r instructors for their thoughtful feedback and encouragement over the course of this project.

\printbibliography

\vspace{12pt}
\end{document}